\documentclass[letterpaper, 10 pt, conference]{ieeeconf}  

\IEEEoverridecommandlockouts                              

\usepackage{graphicx}          
\usepackage{tikz}

\usepackage{amsmath,amssymb,amsfonts,amsthm}
\usepackage{dsfont}
\usepackage{upgreek}
\usepackage{mathtools,mathdots}
\usepackage{nameref}
\usepackage{caption-light}
\usepackage{nicefrac}
\usepackage{enumerate}   
\usepackage{multirow}
\usepackage{multicol}
\usepackage{indentfirst}
\usepackage[
    colorlinks=true,
    allcolors=.
]{hyperref}
\usepackage{balance} 

\DeclareMathOperator*{\argmin}{arg\,min}
			
\newcommand{\Vector}[1]{{#1}}
\newcommand{\Matrix}[1]{\mathrm{#1}}

\newcommand{\defined}{\vcentcolon=}

\newcommand{\reals}{\mathbb{R}}

\newcommand{\continuous}{C}

\newcommand{\Mat}[1][]{\ifthenelse{\equal{#1}{}}{\text{Mat}}{\text{Mat}(#1)}}
\newcommand{\SO}[1]{\textnormal{SO}({#1})}
\newcommand{\SE}[1]{\textnormal{SE}({#1})}
\newcommand{\so}[1]{\mathfrak{so}(#1)}
\newcommand{\se}[1]{\mathfrak{se}(#1)}

\newcommand{\tangent}[1]{
    \ifthenelse{\equal{#1}{}}
    {{T}}
    {{T_{#1}}}
}
\newcommand{\dualtangent}[1]{
    \ifthenelse{\equal{#1}{}}
    {{T^*}}
    {{T_{#1}^*}}
}

\newcommand{\maptoliealgebra}[1]{\left(#1\right)_{\wedge}}

\newcommand{\coadjoint}[1]{
    \ifthenelse{\equal{#1}{}}
    {\textnormal{ad}^*}
    {\textnormal{ad}^*_{#1}}
}

\newcommand{\norm}[1]{\left\lVert#1\right\rVert}
\newcommand{\Frobenius}{\textnormal{F}}

\newcommand{\set}[1]{\left\{#1\right\}}

\newtheorem{remark}{Remark}

\newcommand{\restlength}{L_0}

\newcommand{\position}{x}
\newcommand{\positions}{\Vector{\position}}
\newcommand{\director}{\mathsf{d}}
\newcommand{\orientation}{\Matrix{Q}}
\newcommand{\pose}{\Matrix{q}}
\newcommand{\strain}{\varepsilon}
\newcommand{\strains}{{\Vector{\strain}}}

\newcommand{\curvature}{\kappa}
\newcommand{\curvatures}{\Vector{\curvature}}
\newcommand{\shear}{\nu}
\newcommand{\shears}{\Vector{\shear}}
\newcommand{\internalforce}{n}
\newcommand{\internalforces}{\Vector{\internalforce}}
\newcommand{\internalmoment}{m}
\newcommand{\internalmoments}{\Vector{\internalmoment}}
\newcommand{\internalload}{l}
\newcommand{\internalloads}{\Vector{\internalload}}

\newcommand{\arclengthspace}{\mathbb{S}}

\newcommand{\storedenergy}{W}

\newcommand{\potentialenergy}{\mathsf{U}}

\newcommand{\muscle}{\textnormal{m}}

\newcommand{\LM}[1][]{\textnormal{LM}{\ifthenelse{\equal{#1}{}}{}{_{#1}}}}
\newcommand{\OM}[1][]{\textnormal{OM}{\ifthenelse{\equal{#1}{}}{}{_{#1}}}}

\newcommand{\musclepositions}[1][]{\positions^{\ifthenelse{\equal{#1}{}}{\muscle}{#1}}}
\newcommand{\musclerelativepositions}[1][]{\Vector{\gamma}^{\ifthenelse{\equal{#1}{}}{\muscle}{#1}}}

\newcommand{\musclelength}[1][]{\ell^{\ifthenelse{\equal{#1}{}}{\muscle}{#1}}}
\newcommand{\musclestrain}[1][]{\strain^{\ifthenelse{\equal{#1}{}}{\muscle}{#1}}}
\newcommand{\muscleshears}[1][]{\shears^{\ifthenelse{\equal{#1}{}}{\muscle}{#1}}}
\newcommand{\muscletangent}[1][]{\Vector{\mathsf{t}}^{\ifthenelse{\equal{#1}{}}{\muscle}{#1}}}
\newcommand{\maxmusclestress}[1][]{\sigma^{\ifthenelse{\equal{#1}{}}{\muscle}{#1}}}
\newcommand{\maxmuscleforce}[1][]{\internalforce^{\ifthenelse{\equal{#1}{}}{\muscle}{#1}}_\textnormal{max}}
\newcommand{\muscleforces}[1][]{\internalforces^{\ifthenelse{\equal{#1}{}}{\muscle}{#1}}}
\newcommand{\musclemoments}[1][]{\internalmoments^{\ifthenelse{\equal{#1}{}}{\muscle}{#1}}}
\newcommand{\muscleloads}[1][]{\internalloads^{\ifthenelse{\equal{#1}{}}{\muscle}{#1}}}
\newcommand{\muscleactivation}[1][]{u^{\ifthenelse{\equal{#1}{}}{\muscle}{#1}}}

\newcommand{\staticmuscleactivation}[1][]{\alpha^{\ifthenelse{\equal{#1}{}}{\muscle}{#1}}}

\newcommand{\musclestoredenergy}[1][]{\storedenergy^{\ifthenelse{\equal{#1}{}}{\muscle}{#1}}}
\newcommand{\musclepotentialenergy}[1][]{\potentialenergy^{\ifthenelse{\equal{#1}{}}{\muscle}{#1}}}

\newcommand{\costfunction}{\mathsf{J}}

\newcommand{\inputVector}{X}

\newcommand{\marker}{\textnormal{m}}
\newcommand{\measurementset}{\mathcal{M}}

\newcommand{\basis}{\textnormal{b}}

\newcommand{\BRT}{\textnormal{BR2}}

\DeclareRobustCommand{\IEEEauthorrefmark}[1]{\smash{\textsuperscript{\footnotesize #1}}}
\title{A Neural Network-based Framework for Fast and Smooth\\
Posture Reconstruction of a Soft Continuum Arm}
\author{
    Tixian Wang\IEEEauthorrefmark{\authorrefmark{1},1,2},
    Heng-Sheng Chang\IEEEauthorrefmark{\authorrefmark{1},\authorrefmark{2},1,2},
    Seung Hyun Kim\IEEEauthorrefmark{1},
    Jiamiao Guo\IEEEauthorrefmark{1},
    Ugur Akcal\IEEEauthorrefmark{2,3},
    Benjamin Walt\IEEEauthorrefmark{1},\\
    Darren Biskup\IEEEauthorrefmark{1},
    Udit Halder\IEEEauthorrefmark{5},
    Girish Krishnan\IEEEauthorrefmark{4},
    Girish Chowdhary\IEEEauthorrefmark{2,3},
    Mattia Gazzola\IEEEauthorrefmark{1},
    Prashant G. Mehta\IEEEauthorrefmark{1,2}
    \thanks{This work was financially supported by ONR MURI (N00014-19-1-2373), AFOSR award (FA9550-23-1-0060), and NSF award (2336137).}
    \thanks{This work was carried out in part in the Center for Autonomy Robotics Laboratories at the University of Illinois Urbana-Champaign.}
}

\begin{document}
  \maketitle
    \bstctlcite{IEEEexample:BSTcontrol}

  \let\thefootnote\relax\footnote{
    \authorrefmark{1}These authors contributed equally.
    \IEEEauthorrefmark{\scriptsize 1}Mechanical Science and Engineering,
    \IEEEauthorrefmark{\scriptsize 2}Coordinated Science Laboratory,
    \IEEEauthorrefmark{\scriptsize 3}Field Robotics Engineering and Sciences Hub (FRESH),
    \IEEEauthorrefmark{\scriptsize 4}Industrial and Enterprise Systems Engineering,
    University of Illinois Urbana-Champaign. 
    \IEEEauthorrefmark{\scriptsize 5}Mechanical Engineering, University of South Florida.
    \authorrefmark{2}C.A.: hschang2@illinois.edu. 
    GitHub repo: \url{https://github.com/tixianw/FastRodReconstruction}
  }

  \begin{abstract}
    A neural network-based framework is developed and experimentally demonstrated for the problem of estimating the shape of a soft continuum arm (SCA) from noisy measurements of the pose at a finite number of locations along the length of the arm.  
    The neural network takes as input these measurements and produces as output a finite-dimensional approximation of the strain, which is further used to reconstruct the infinite-dimensional smooth posture. 
    This problem is important for various soft robotic applications. 
    It is challenging due to the flexible aspects that lead to the infinite-dimensional reconstruction problem for the continuous posture and strains. 
    Because of this, past solutions to this problem are computationally intensive. 
    The proposed fast smooth reconstruction method is shown to be five orders of magnitude faster while having comparable accuracy.
    The framework is evaluated on two testbeds: a simulated octopus muscular arm and a physical BR2 pneumatic soft manipulator.

  \end{abstract}

  \section{Introduction} \label{sec:intro}
    
    Soft manipulators have gained increasing attention from the robotics community because of their unique capabilities and advantages over traditional rigid manipulators~\cite{zongxing2020research,chen2022review}. 
    These systems, that are often bio-inspired from animals such as elephants and octopuses~\cite{leanza2024elephant,laschi2012soft}, are characterized by their intrinsic compliance, dexterity, and adaptability, enabling them to interact with environment safely and handle delicate objects~\cite{venter2017self,banerjee2018soft,chen2021soft}. 
    Typically made from highly deformable materials such as elastomers, soft manipulators can perform complex, continuous deformations along their entire length including bending, twisting, and elongation. 

    These advantageous properties of soft robotic manipulators, however, also present significant challenges in the applications involving control~\cite{gerboni2017feedback,della2020model}, path planning~\cite{singh2021curve}, force estimation~\cite{mei2024simultaneous}, and human-robot interaction~\cite{ku2024soft}. 
    In practical settings, many of these applications require an estimation of the (continuum) shape of the soft robot. 
    In contrast to the rigid counterparts, the shape of a soft robot has infinite degrees of freedom. 
    This makes the problem challenging and the solutions computationally intensive, requiring significant computing resources in real-time applications.  
    
    \begin{figure}[t]
      \vspace{15pt}
      \centering
      \includegraphics[width=0.9\columnwidth, trim = {0pt 0pt 0pt 0pt}, clip = false]{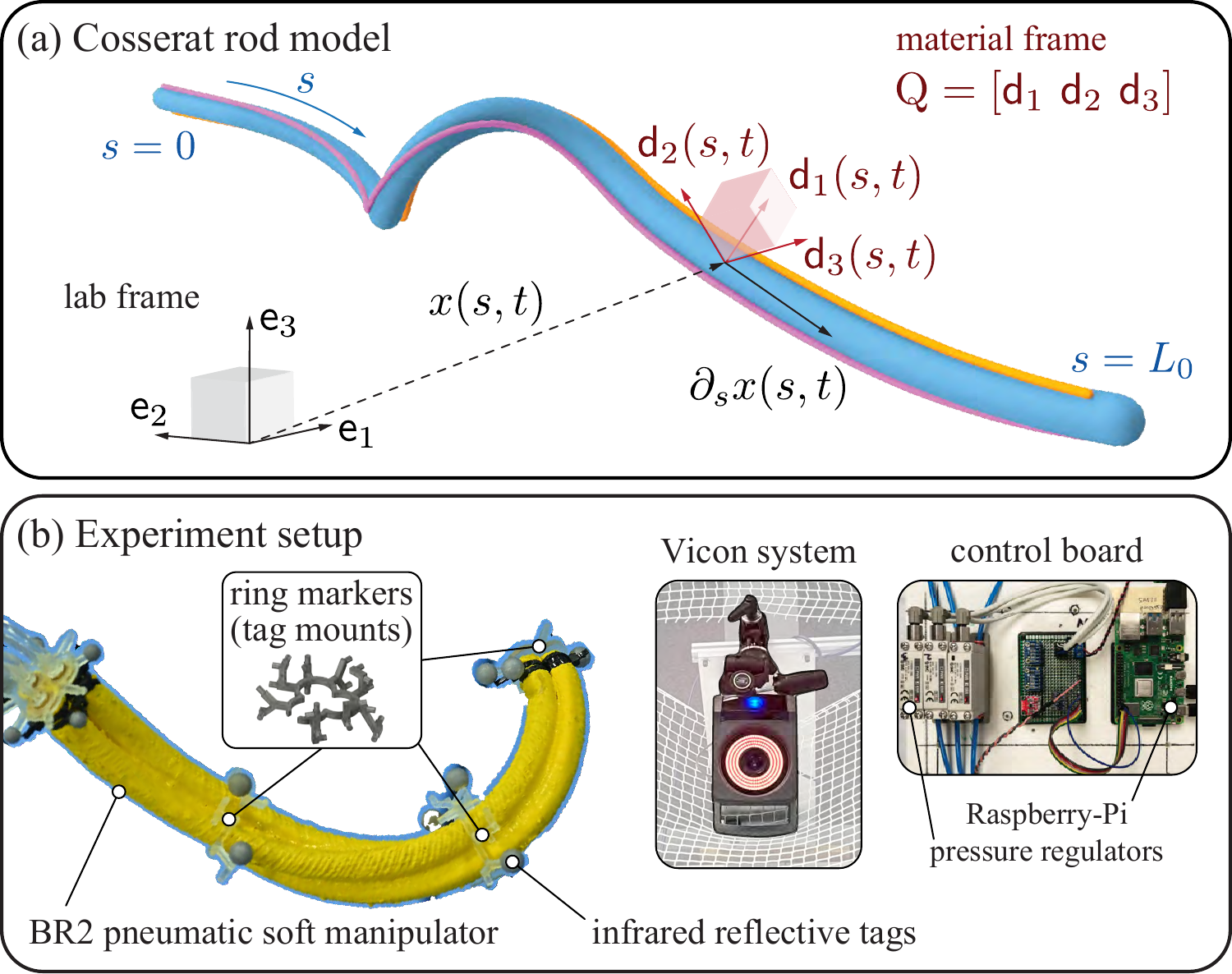}
    	\caption{
              (a) Cosserat rod model for a slender soft continuum arm. 
              (b) experiment setup: a $\BRT$ pneumatic soft manipulator is set up in the Vicon motion capture arena. 
              Actuation signals are generated from Raspberry-Pi 4 and relayed to SMC pressure regulators.
              Ring markers are designed to hold the infrared reflective tags for the Vicon system to track.}
        \label{fig:exp_setup}
  	\vspace{-15pt}
    \end{figure}

    Research on shape estimation for soft robots has grown in recent years, focusing on two main approaches: proprioception and exteroceptive sensing.
    Proprioceptive methods use on-board sensors like Fiber Bragg Grating~\cite{xu2016curvature,zhang2018shape}, electromagnetic sensor~\cite{bergamini2014estimating,guzman2020low}, capacitive flex sensor~\cite{toshimitsu2021sopra}, or liquid metal sensors~\cite{park2013smart,kim2019soft}. 
    While effective for real-time reconstruction, these can be invasive and may alter the material's elastic properties and interfere with the soft robot functionality. 
    Additionally, they may require calibration with the ground truth provided from the exteroceptive sensing~\cite{truby2020distributed,wang2024shape}.
    

     Exteroceptive perception leverages vision systems to reconstruct the shapes of the soft manipulators. 
     The vision-based approaches can be categorized into two types: marker-based and markerless methods. 
     The former uses point extraction methods (e.g., direct linear transformation~\cite{abdel2015direct}, optical flow~\cite{kim2022physics}) to localize reflective markers and then interpolate discrete points into continuous shape, from which local strains can also be estimated. 
     On the other hand, markerless methods typically extract the soft manipulator shape from the images and then perform reconstruction, such as skeletonization~\cite{fan2020image} and deep learning~\cite{zou2022deep}.

    In real-world scenarios, time-sensitive tasks like sensory feedback control \cite{della2018dynamic,wang2022sensory, wang2024neural}, require real-time shape estimation. 
    Many of the vision-based methods above rely on post-processing of the data or the algorithm requires heavy computation, which are not suitable for real-time tasks. 
    Some marker-based methods use parameterized strain functions to increase the interpolation speed from simple curvature model like piece-wise constant curvature~\cite{hannan2005real} to more complex representation such as cubic spline method~\cite{godage2018real} and multi-segment PH curves~\cite{bezawada2022shape}. 
    These approaches are limited to specific systems and applications. 
    Another way to achieve real-time sensing is through deep learning which typically uses images as input data and train a convolutional neural network~\cite{hofer2021vision,rong2024vision,zheng2024vision}. 
    These methods require ground truth data in order to perform supervised learning. 
    However, it could be expensive and challenging to obtain extensive training data with physical soft robots. 

    This paper aims to provide a framework to perform real-time smooth posture reconstruction as well as continuous strain estimation for a generalized soft continuum arm given a finite number of poses collected from vision system.

    \subsection*{Contributions}

      This paper builds on a body of work from our group \cite{chang2023energy, kim2022physics, shih2023hierarchical} on control-oriented modeling, estimation, and learning of soft continuum arms. 
      The unique contributions of this paper are as follows:

      \begin{enumerate}
          \item A neural network-based framework for fast and smooth posture reconstruction with strain estimation through unsupervised learning.
          \item Simulation demonstration of the framework for a computational model of an octopus muscular arm. 
          \item Robotic application of the framework through the physical $\BRT$ pneumatic soft manipulator operating in a  Vicon motion capture environment.
      \end{enumerate}

      The remainder of this paper is organized as follows. 
      The model of soft continuum arm system and the problem formulation are introduced in Sec.~\ref{sec:problem_formulation}. 
      The proposed neural network framework and the fast smooth reconstruction method are described in Sec.~\ref{sec:solution}. 
      Performance of the method is demonstrated with both simulation and robotic experiments in Sec.~\ref{sec:results}. 
      Conclusion and future work appear in Sec.~\ref{sec:conclusion}.

  \section{Problem Formulation} \label{sec:problem_formulation}

    
    \subsection{Mathematical model of soft continuum arm}
      A soft continuum arm (SCA) robot depicted in Fig.~\ref{fig:exp_setup} is modeled as a special Cosserat rod \cite{antman1995nonlinear}.
      The arm is fixed at its base with nominal length $\restlength$.
      The arc-length parameter along its center-line is denoted as $s\in\arclengthspace=[0,\restlength]$. 
      The base is at $s=0$ and the tip is at $s=\restlength$.
      With respect to an inertial lab frame $\set{\mathsf{e}_i}_{i=1}^3$, the \emph{posture} of the arm is written as
      \begin{equation}
        \pose(s)\defined\begin{bmatrix}
          \orientation(s) & \positions(s)\\
          0 & 1
        \end{bmatrix}\in\SE{3},\quad \forall~s\in\arclengthspace
      \end{equation}
      where $\SE{3}$ is the three-dimensional special Euclidean Lie group.
      The center-line \emph{position} and \emph{orientation} are denoted as $\positions(s)\in\reals^3$ and $\orientation(s)=[\director_1(s)~\director_2(s)~\director_3(s)]\in\SO{3}$, respectively.
      The orthonormal \emph{directors} $\set{\director_i(s)}_{i=1}^3$ define the \emph{material frame} at $s\in\arclengthspace$.


      The \emph{kinematic equation} of the arm is given by,
      \begin{equation}\label{eq:spatial_transformation}
          \partial_s\pose(s)=\pose(s)\strains(s),\quad \forall~s\in\arclengthspace,\quad \text{with }\pose(0)=\pose^{(0)}
      \end{equation}
      where $\pose^{(0)}\in\SE{3}$ is the base of the posture, and $\strains(s)\in\se{3}$ is the \emph{strain} at $s\in\arclengthspace$ along the soft arm.
      The vector space $\se{3}$ is the Lie algebra associated with the Lie group $\SE{3}$.
      The strain $\strains$ takes the form 
      \begin{equation}
          \strains\defined\begin{bmatrix}
              \maptoliealgebra{\curvatures} & \shears\\
              0 & 0
          \end{bmatrix},\quad\curvatures\defined\begin{bmatrix}
              \strains^1 \\ \strains^2 \\ \strains^3
          \end{bmatrix},\quad\shears\defined\begin{bmatrix}
              \strains^4 \\ \strains^5 \\ \strains^6
          \end{bmatrix}
      \end{equation}
      where $\curvatures(s)\in\reals^3$ are the three angular strains and $\shears(s)\in\reals^3$ are the three linear strains.
      The angular strains $\strains^1, \strains^2$ are \emph{curvatures} along axis $\director_1, \director_2$, and $\strains^3$ is the \emph{twist} along $\director_3$.
      The linear strains $\strains^4, \strains^5$ are \emph{shears} along axis $\director_1, \director_2$, and $\strains^6$ is the \emph{stretch} along $\director_3$. 
      The hat map $\maptoliealgebra{\cdot}$ is the isomorphic operator that maps a vector in $\reals^3$ into an element in $\so{3}$, the Lie algebra associated with the Lie group $\SO{3}$.

      The dynamics of the soft continuum arm are modeled using a Hamiltonian control system, which is numerically implemented in \emph{Elastica} \cite{gazzola2018forward} with two simulators:

      \begin{enumerate}
        \item Simulator for an octopus arm: The control input is from the internal musculature of the arm and results in internal muscular contraction forces. Details of which can be found in~\cite{chang2023energy} with the repository COOMM \cite{chang2023coomm}. 
        \item Simulator for an SCA robot $\BRT$: The control input models the effect of pneumatic bending and twisting actuation, the details of which can be found in the repository COBRA \cite{chang2024cobra}. 
        This model is part of an ongoing research project with prior work reported in \cite{Krishnan2012FREEdeformation, Uppalapati2021BR2, Krishnan2015FREEdeformation}.
        
      \end{enumerate}
      


    \begin{figure*}[!t]
        \centering
        \includegraphics[width=0.9\textwidth, trim = {0pt 0pt 0pt 0pt}]{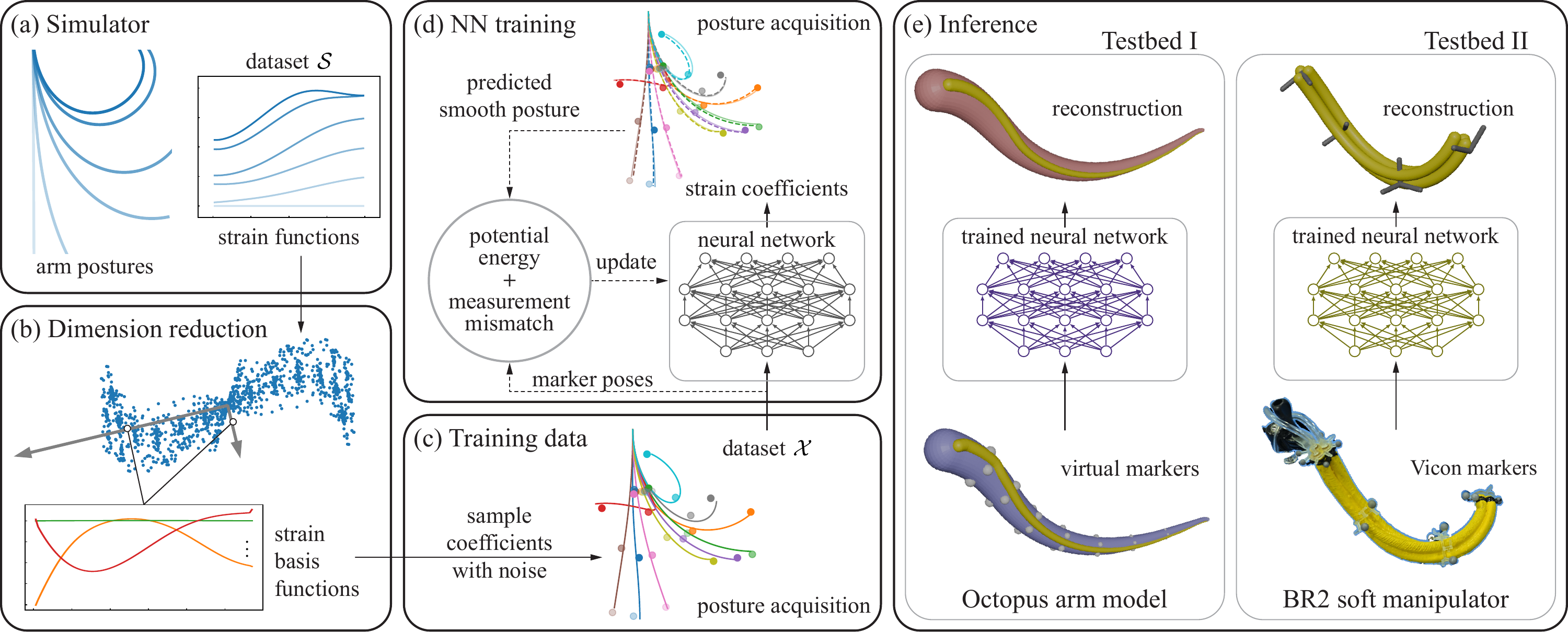}
        \caption{
              Pipeline of the proposed fast smooth reconstruction method: 
              (a) A dynamic simulator is used to generate a initial dataset $\mathcal{S}$ of soft arm deformation trajectories in a suitable workspace. 
              (b) Principle component analysis is performed on the strain functions in dataset $\mathcal{S}$. Each type of strain is represented by a finite number of basis functions. 
              (c) Training data $\mathcal{X}$ contains marker poses obtained from strain coefficient sampling and posture acquisition.
              (d) At the training stage, a multilayer perceptron neural network model is learnt in an unsupervised manner to minimize the phyiscs-informed objective function.
              (e) Inference stage is carried out on two testbeds: a computational octopus muscular arm and a physical $\BRT$ pneumatic soft manipulator.
            }
        \label{fig:fast_NN_pipeline}
  	\vspace{-12pt}
    \end{figure*}

    \subsection{Reconstruction problem and prior work}\label{sec:reconstruction_problem_formulation}


      To help motivate the reconstruction problem, the guiding example is the experimental $\BRT$ robot arm whose shape needs to be tracked with cameras. 
      For this purpose, $N_\marker$ markers are placed at known fixed arc-length locations $\set{s_m}_{m=1}^{N_\marker}$ along the arm, with $0=s_0<s_1<\cdots<s_{N_\marker}=\restlength$. 
      A control input is applied which results in the motion of the arm.
      
      For each fixed time, the \emph{pose} at each marker location $s_m$ is measured with multiple cameras.
      The measured pose is denoted as $\pose^{(m)}\in\SE{3}$ and modeled as a noisy measurement of the true pose $\pose(s_m)$. 
      Together with the marker locations, the measurement set is denoted as $\measurementset=\set{(s_m,\pose^{(m)})}_{m=1}^{N_\marker}$.
      
      The objective of \emph{fast smooth reconstruction} is to compute a smooth arm posture $\hat{\pose}(s)\in\SE{3}~\forall~s\in\arclengthspace$ based on the measurement set $\measurementset$ for each time frame.

      To solve this reconstruction problem, the following optimization problem was introduced in our prior work~\cite{kim2022physics}. 
      
      \noindent \textbf{Optimization problem.}
      Let the admissible strain $\strains$ be the decision variables, and the objective function as
      \begin{equation}\label{eq:objective}
        \costfunction(\pose;\measurementset) \defined \potentialenergy(\pose) + \frac{\eta}{2}\Phi(\pose; \measurementset)
      \end{equation}
      where $\potentialenergy$ is the potential energy of the arm, $\eta>0$ is a regularization parameter, and $\Phi$ is the measurement mismatch cost defined as follows
      \begin{equation*}
        \Phi(\pose;\measurementset)\defined\sum_{m=1}^{N_\marker}\frac{\norm{x(s_m)-x^{(m)}}^2}{(\restlength)^2}+\frac{\norm{\orientation(s_m)-\orientation^{(m)}}_{\Frobenius}^2}{8}
      \end{equation*}
      with $\norm{\cdot}_{\Frobenius}$ as the Frobenius norm.
      The optimization problem reads as follows:
      \begin{equation}
        \hat{\strains}=\argmin_{\strains}~\costfunction(\pose;\measurementset)\quad\text{s.t.} ~  \text{kinematic constraint~\eqref{eq:spatial_transformation}}
        \label{eq:optimization_problem}
      \end{equation}
      The reconstructed posture $\hat{\pose}$ is then obtained by integrating the strain $\hat{\strains}$ through the kinematic equation~\eqref{eq:spatial_transformation}. 

      \begin{remark}
        The proposed optimization problem can be regarded as quasi-static formulation whereby the optimization problem is solved at each fixed time-step. 
        The quasi-static setup is motivated from the work of \cite{albeladi2021vision, fu2021continuous} who introduced the following optimization problem
          \begin{equation}\label{eq:raw_reconstruction_problem}
              \min_{\strains}~\potentialenergy\quad\textnormal{s.t.}\quad\partial_s\pose=\pose\strains,~\textnormal{with }\pose(s_m)=\pose^{(m)}~\forall~m
          \end{equation}
          While such an approach offers several benefits, including stability and uniqueness \cite{bretl2014quasi,till2017elastic}, it typically results in discontinuous strains. 
          The optimization problem introduced in our prior work \cite{kim2022physics} incorporates the measurement set $\measurementset$ in a global manner with the decision variable set as the strain rate.
          A key advantage is that it yields a continuous solution for both posture and strain. 
          The continuity property is crucial for accurately capturing the smooth, natural deformation of the flexible arm and provides a realistic and physically valid reconstruction compared to piecewise continuous methods.    
      \end{remark}

      A key limitation of our previous work is that the optimization problem is computationally expensive. 
      The algorithm proposed in \cite{kim2022physics} relies on a forward-backward integration: the forward integral for the state (pose) and backward integral for the costate (internal loads). 
      At each iteration, the strain is updated till convergence. 
      While the results are promising on the evaluated benchmarks, its computational intensity, often requiring thousands of iterations per frame, makes it impractical for real-time applications.
      This limitation motivates our current work, which aims to develop a more efficient solution while maintaining accuracy.
      This is the main contribution of our work and it is detailed in the next section.

  \section{Fast \& Smooth Reconstruction Framework} \label{sec:solution}

  \begin{figure*}[!t]
  	\centering
  	\includegraphics[width=0.9\textwidth, trim = {0pt 0pt 0pt 0pt}]{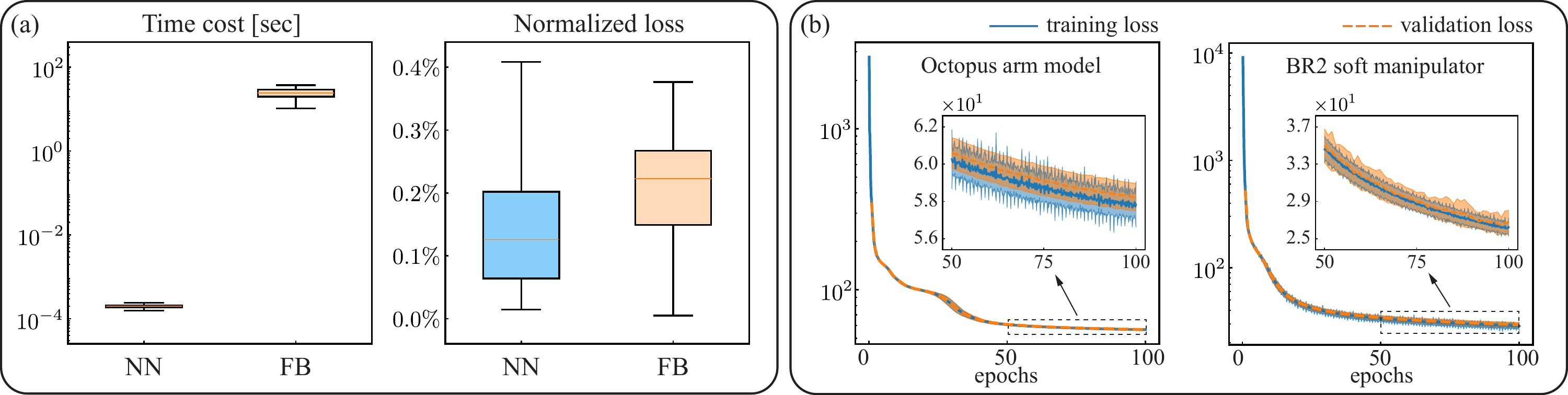}
  	\caption{
          (a) Performance comparison of the proposed reconstruction method (NN) with the benchmark forward-backward algorithm (FB) in \cite{kim2022physics}. 
          Left column reports the computational time cost for one reconstruction sample. 
          NN provides smooth posture reconstruction over five orders of magnitude faster than FB. 
          Right column shows the normalized loss which indicates that NN achieves the reconstruction accuracy in a comparable scale to FB. 
          The normalized loss is the loss~\eqref{eq:objective} normalized by the regularization parameter $\eta$ and the number of markers $N_\marker$. 
          (b) Training and validation losses over epochs for two testbeds, a computational octopus muscular arm and a physical $\BRT$ pneumatic soft manipulator. 
          Each plot shows the mean and standard deviation of the training and validation losses over a total of fifty independent training experiments. Insets show the change of losses after 50 epochs.}
  	\label{fig:benchmark_training}
  	\vspace{-12pt}
  \end{figure*}



    The key insight driving the present work is that the solution of the constrained optimization problem~\eqref{eq:optimization_problem} defines a map from the measurements $\measurementset\in(\arclengthspace\times\SE{3})^{N_\marker}$ to the strains $\strains \in \continuous(\arclengthspace;\se{3})$. 
    The idea of this paper is to learn this map using a neural network (NN). 
    Once a neural network is learned, it can be used in real-time operations to convert the marker poses (input of NN) to the estimated strains (output of NN). 
    Once the strains are known, the kinematic equation~\eqref{eq:spatial_transformation} can be used to recover the posture.      
    
    While the idea is simple, its application to the problem is far from straightforward. 
    The main difficulties are as follows:
    \begin{enumerate}
        \item The solution to the optimization problem \eqref{eq:optimization_problem} is an infinite-dimensional strain function (an element of $\continuous(\arclengthspace;\se{3})$), which is an issue for a neural network whose output is always finite-dimensional.
        \item While the input (marker poses $\measurementset$) to the neural network is available, the output (strain $\strains$) is not so. 
        In principle, one may use the forward-backward algorithm to obtain an estimated strain as labeled output, but that would be computationally expensive.
        \item The experiments are costly requiring the use of expensive high speed marker tracking facilities. 
        This limits the amount of marker pose measurements (as training data) that can be obtained from running the experiments.
    \end{enumerate}

    To address these challenges, a systematic procedure is designed and depicted in Fig.~\ref{fig:fast_NN_pipeline}. 
    The key steps of the procedure are as follows:

    \subsection{Simulator}
      To overcome the challenge of costly experiments, a simulation framework \emph{Elastica} is used to obtain a small initial dataset of strain trajectories for the training of our NN-based fast smooth reconstruction.
      We run simulations based on the Hamiltonian control system with various control inputs, and record the resulting strain trajectories as the initial dataset $\mathcal{S}=\set{\strains_t}$. 
      These simulations capture time-evolving postures and all six strain modes along the arm.
      It serves as the foundation for subsequent dimension reduction and guides the creation of a larger set of training data.

      
      

  \begin{figure*}[!t]
  	\centering
  	\includegraphics[width=0.9\textwidth, trim = {0pt 0pt 0pt 0pt}]{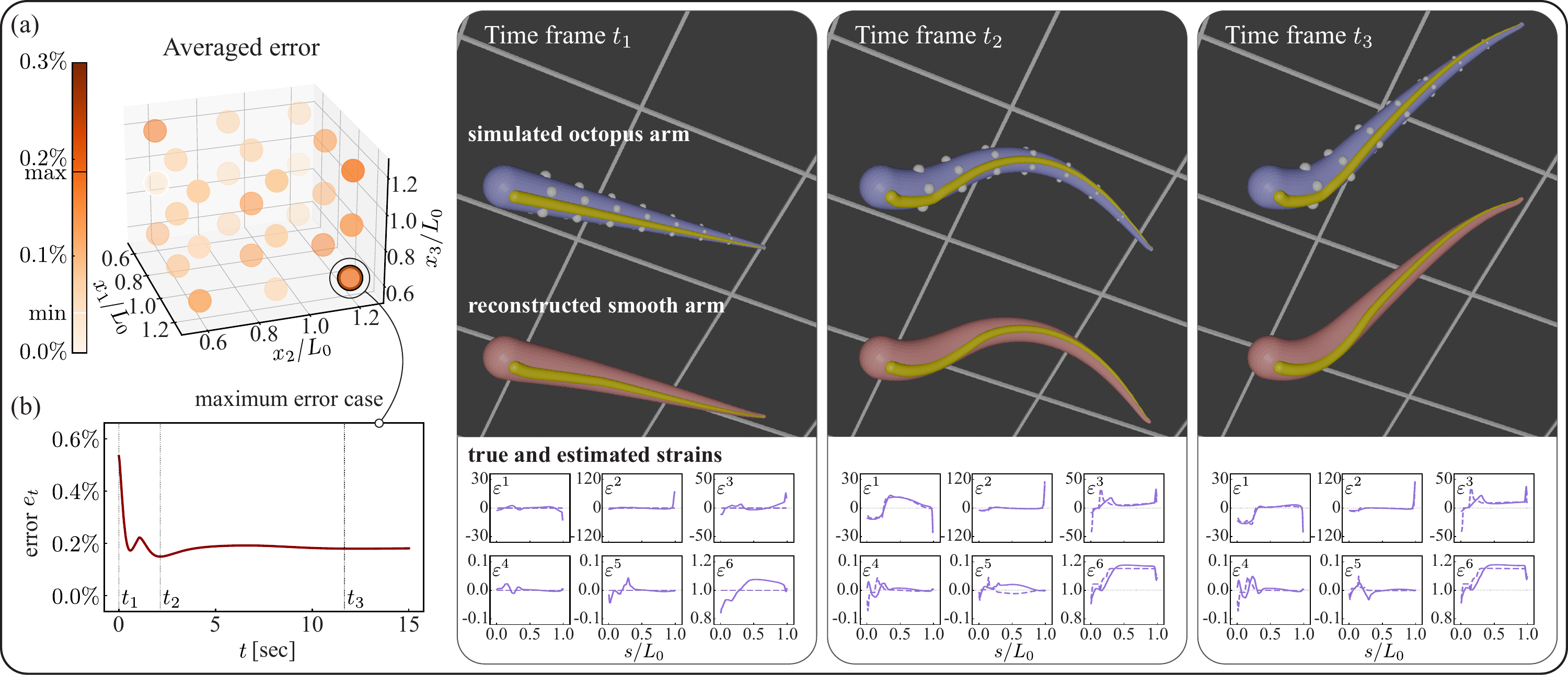}
  	\caption{
          Performance of the proposed framework on the simulated octopus muscular arm model:  
          (a) Averaged arm reconstruction error over samples for 27 targets uniformly distributed in the workspace $\mathcal{W}$. 
          The target with maximum (minimum) averaged error is marked with black (white) circle, and on the color bar as well.
          (b) The target with the maximum averaged error from part (a) is chosen for plotting the arm reconstruction error $e_t$ over time. 
          The three panels on the right demonstrate the simulated octopus arm postures at three time frames $t_1,t_2,t_3$ with their corresponding reconstructed arm postures. 
          For each time frame, all six estimated strains $\hat{\strains}_t(s)$ (in solid lines) are compared with the ground truth $\strains_t(s)$ (in dashed lines) from octopus arm simulator.
        }
  	\label{fig:octopus_reconstruction_result}
  	\vspace{-12pt}
  \end{figure*}
    
    \subsection{Dimension reduction}

      The aim of the dimension reduction is to approximate the six strains via a finite number of basis functions as follows
      \begin{equation}\label{eq:finite_dimensional_representation}
        \strains^i(s) = \sum_{j=0}^{N_\basis} \alpha^i_j \bar{\strains}^i_j(s),\quad \forall~i=1,2,\ldots,6,~\forall s\in\arclengthspace
      \end{equation}
      where for each strain $i$, $N_\basis$ is the number of basis functions, $\bar{\strains}^i_j(s)$ is the $j^\text{th}$ basis, and $\alpha^i_j$ is the corresponding coefficient. 
      The basis functions are obtained via the principal component analysis (PCA) \cite{abdi2010principal} on the standardized initial dataset $\mathcal{S}$ (with mean and variance functions).
      For the $i^\text{th}$ strain, $\bar{\strains}^i_0(s)$ is the mean function with $\alpha^i_0=1$, and the largest $N_\basis$ eigenvalues are chosen with the basis functions $\bar{\strains}^i_j(s)$ set as eigenfunctions weighted with the variance function.
      The coefficients $\alpha^i_j$ serve two purposes:
      \begin{enumerate}
          \item Instead of infinite dimensional strain functions, strain coefficients are now the output of the neural network. 
          \item The distribution of coefficients are used to generate training data via sampling.
      \end{enumerate}

    \subsection{Training data}

      The process of generating the training data for the neural network involves the following steps:
      \begin{enumerate}
          \item \emph{Sampling}: For each coefficient $\alpha^i_j$, $K$ samples are generated based on the Gaussian distribution with mean and variance which are empirically obtained based on the PCA results. 
          \item \emph{Posture acquisition}: Based on the sampled coefficients, the strain functions $\{\strain_k(s)\}_{k=1}^K$ are obtained through~\eqref{eq:finite_dimensional_representation}. The sampled strains are further integrated to obtain postures $\{\pose_k(s)\}_{k=1}^K$ by the kinematic equation~\eqref{eq:spatial_transformation}.
          \item \emph{Training data}: For each posture sample, we measure $N_\marker$ marker poses $\{\pose^{(m)}_k\}_{m=1}^{N_\marker}$ at locations $\set{s_m}_{m=1}^{N_\marker}$ along the center-line with added Gaussian noise to simulate measurement uncertainty.
          The collection of noisy marker poses forms the comprehensive training data $\mathcal{X}=\{\{s_m,\pose_k^{(m)}\}_{m=1}^{N_\marker}\}_{k=1}^K$.
      \end{enumerate}
      This approach allows us to expand our initial small dataset $\mathcal{S}$ into a larger, more diverse training data $\mathcal{X}$ that captures the underlying distributions of arm postures and strains, while also accounting for potential measurement noise in real-world scenarios.
      The noise level can be adjusted to reflect the expected error in the physical measurement system.

        

    \subsection{Neural network training}
      
      A Multilayer Perceptron (MLP) or a feedforward neural network is designed to learn the mapping from marker poses to coefficients of strain basis functions. Details are as follows:
      \begin{enumerate}
          \item \emph{Input of NN}: The input layer comprises each marker's position and two directors \cite{zhou2019continuity} (the rest one can be derived from orthonormality), represented as:
              \begin{equation}
                  \inputVector^{(m)} = [
                      \positions^{(m)}~\director^{(m)}_1~\director^{(m)}_3
                ],~\forall~m=1,2,\dots,N_\marker
              \end{equation}
              The number of entries in each $X^{(m)}$ is 9.
        \item \emph{Output of NN}: The output layer nodes represent $\alpha^i_j$, which are the coefficients of all strain basis functions.
        \item \emph{NN architecture}: 
          The NN model consists of two hidden layers with Sigmoid Linear Unit (SiLU) activation functions.
          The number of hidden neurons are adjusted based on the number of markers (input size) and number of basis functions (output size). 
        \item \emph{Training setting}: 
          For data preparation, the training data $\mathcal{X}$ is randomly split into training (80\%) and validation (20\%) sets. 
          The loss function follows the objective function~\eqref{eq:objective} with the regularization parameter $\eta=10^4$.
          The model is trained for 100 epochs using the Adam optimizer with learning rate $10^{-3}$ and batch size 128.
        \item \emph{Training method}: 
          In each epoch, training set is shuffled into mini-batches to perform forward path, compute loss, conduct back propagation, and update weights. 
          After each epoch, the model is evaluated with validation set to monitor performance and prevent overfitting.
      \end{enumerate}

      The choice of loss function allows for training without labeled data, eliminating the need for ground truth strain measurements.
      The key advantages of this unsupervised learning approach are that the model is trained with the loss that is physics informed, and the predicted posture not only minimizes the potential energy $\potentialenergy$ but also assimilates the measurement set $\measurementset$.

      
    

    \subsection{Inference}
      During the inference stage, the trained model is deployed to reconstruct the arm posture and strains for each time frame $t$ independently. 
      The process begins by transforming the marker pose data from $\measurementset_t$ into the required input format $X_t=\{X^{(m)}_t\}_{m=1}^{N_\marker}$ for the model. 
      These processed marker poses are then fed into the model, which outputs the coefficients for the strain basis functions. 
      Using these inferred coefficients, strain functions $\hat{\strains}_t$ are estimated along the entire length of the arm. 
      Subsequently, a reconstructed posture $\hat{\pose}_t$ is obtained by integrating the kinematic equation \eqref{eq:spatial_transformation}.
      This approach generates a smooth, physically consistent representation of the arm posture and strains for each time frame in real-time.
      A comparison with the benchmark algorithm in \cite{kim2022physics} on the performances of reconstruction time cost and objective function loss is depicted in Fig.~\ref{fig:benchmark_training}(a).
      All training and inference computations were executed on a laptop with an Intel Core i7-12700H processor (2.3GHz, 6P+8E cores), 16GB 3200MHz DDR4 memory, and no dedicated GPU.

\section{Results and Analysis} \label{sec:results}

In this paper, we evaluate the proposed fast smooth reconstruction method in two testbeds. The first is the simulation demonstration of a computational octopus muscular arm. The second is the robotic experiment for a pneumatic soft manipulator called $\BRT$. In the rest of the section, the setup and training preparation are presented for each of the two testbeds, followed by the discussion of the training and inference results.
  
\subsection{Testbed I: computational octopus muscular arm} \label{sec:simulation}

  \begin{figure*}[!t]
  	\centering
  	\includegraphics[width=0.9\textwidth, trim = {0pt 0pt 0pt 0pt}]{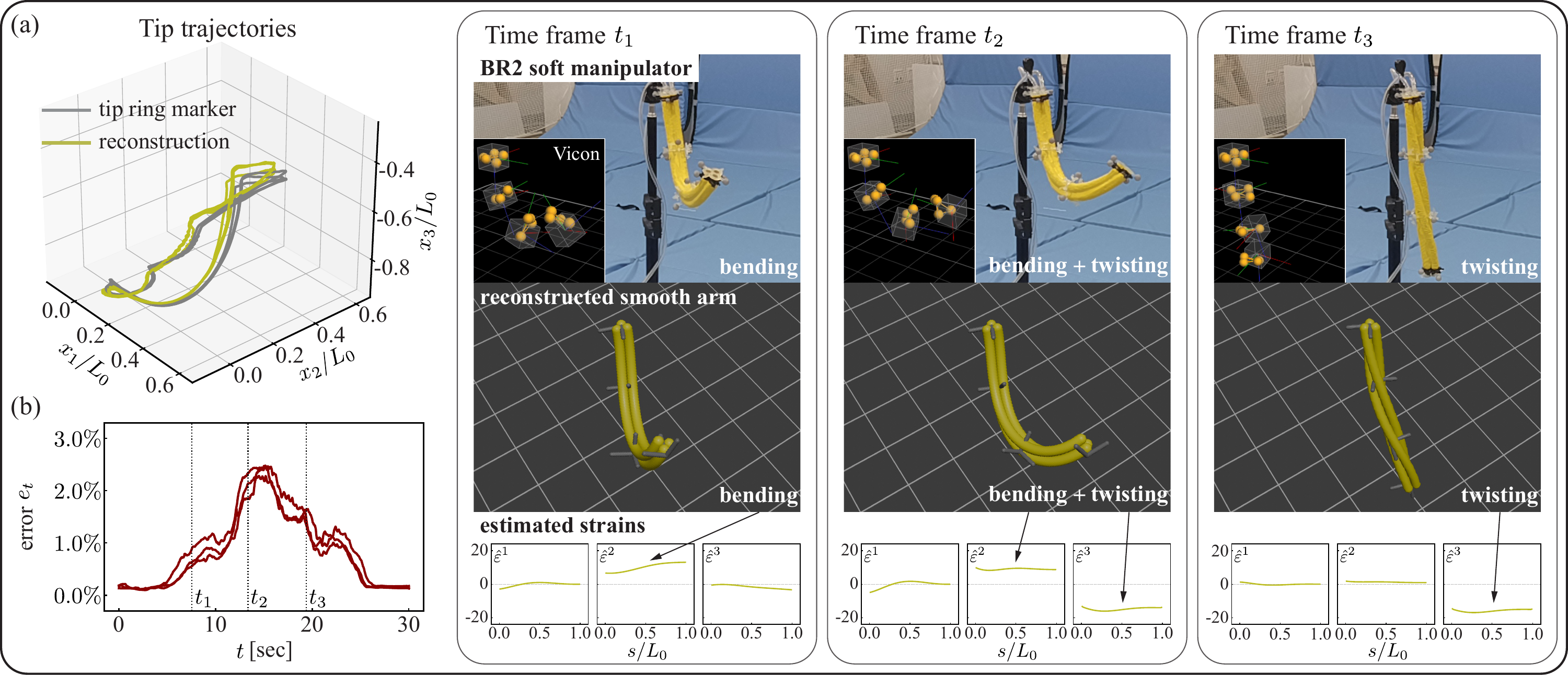}
  	\caption{
          Performance of the proposed framework on the physical $\BRT$ soft manipulator: 
          (a) Tip trajectories of the reconstructed $\BRT$ and the physical $\BRT$ (tip ring marker) from the repeated motion cycles. 
          (b) The arm reconstruction error $e_t$ over time. 
          The error curves are overlapped for each cycle. 
          The three panels on the right demonstrate the camera frames of $\BRT$ soft manipulator, the poses from Vicon ring markers, and corresponding reconstructed smooth arm postures. 
          Measured marker poses are rendered in grey directors to illustrate the reconstruction accuracy. 
          For each time frame, all three estimated curvatures (angular strains) are demonstrated, and represent the pure bending, bending-twisting combination, and pure twisting.}
  	\label{fig:BR2_reconstruction_result}
  	\vspace{-12pt}
  \end{figure*}

\noindent
\textbf{Simulation setup.}
The small dataset $\mathcal{S}$ is collected using the COOMM simulator, an octopus muscular arm model. 
A tapered arm of length $\restlength=0.2$ [m] is initialized straight and the energy shaping control is applied to perform the target reaching tasks~\cite{chang2023energy, chang2020energy, chang2021controlling}. 
A total of 27 numerical experiments are simulated, with targets uniformly distributed throught out the workspace $\mathcal{W}=[0.6\restlength,1.2\restlength]^3$.
A number of $N_\marker=8$ virtual markers are evenly spaced along the arm.

\noindent
\textbf{Dimension reduction.}
A number of $N_\basis=4$ top components from PCA for each strain is selected with their coefficient distributions used for generating the training data $\mathcal{X}$.

\noindent
\textbf{Neural network training.}
The input and output size of the neural network are $9N_\marker=72$ and $6N_\basis=24$. 
The number of neurons in the hidden layers are chosen to be 128 and 64.
The training and validation losses over 100 epochs for the octopus arm model are depicted in the left column of Fig.~\ref{fig:benchmark_training}(b). 
Both losses decrease rapidly within the first 50 epochs, after which the learning starts to converge. 
We pick the best out of fifty trained models at the $100^\text{th}$ epoch for inference. 



\noindent
\textbf{Inference.} 
For every simulation time step $t$, the input $X_t$ is extracted from the marker poses $\measurementset_t$. 
The reconstructed posture $\hat{\pose}_t$ is obtained as the output of the framework.
Fig.~\ref{fig:octopus_reconstruction_result}(a) shows the inference performance for arm reaching motions towards 27 targets in the workspace $\mathcal{W}$. 
The performance metric for each target is the averaged reconstruction error over time.
The reconstruction error at time $t$ is defined as $e_t\defined\frac{1}{N_\marker}\Phi(\hat{\pose}_t;\measurementset_t)$ which is the measurement mismatch cost per marker. 
The results show a relatively uniform performance among all targets and all the averaged reconstruction errors fall below 0.2\% level.
The case with maximum averaged reconstruction error is further reported in Fig.~\ref{fig:octopus_reconstruction_result}(b). 
At three time frames (highest error, lowest error, and final), the simulated octopus arm posture with the corresponding reconstructed arm and estimated strains are presented in the right three panels of Fig.~\ref{fig:octopus_reconstruction_result}. 
The comparison shows that the reconstruction matches well with the true arm posture and we have a good estimation for all six strains.


\subsection{Testbed II: physical \BRT\ pneumatic soft manipulator} \label{sec:experiment}


\noindent
\textbf{Experiment setup.}
While our framework applies to various SCA systems, here we focus on the $\BRT$ pneumatic soft manipulator \cite{Uppalapati2021BR2}. 
It takes three pressure inputs and performs bending and two twisting (clockwise and counterclockwise rotations) motions, hence the name $\BRT$. 
The combination of motions creates a variety of representative 3D deformations. 
The $\BRT$ is of length $\restlength=0.3$ [m] along which a number of $N_\marker=3$ ring markers are evenly placed.
The $\BRT$ is set inside the Vicon motion capture arena.
The Vicon system \cite{merriaux2017study}, widely employed in various disciplines including biomechanics and robotics \cite{pfister2014comparative, bauer2024an}, functions by detecting the reflective tags or markers (groups of tags) with a network of motion cameras.
To receive the marker pose message, a custom ROS 2 Foxy Docker image \cite{rostwo2022, chang2024Vicon, camisavicon2020} is used.
A complementary filter on $\SE{3}$ \cite{mahony2005complementary} is implemented for the stability of marker pose reading.



\noindent
\textbf{Simulation setup.}
The dataset $\mathcal{S}$ is gathered using COBRA, a simulator for $\BRT$.
A combination of various pressure inputs of bending and twisting control are used to generate $\BRT$ motions where the angular strain data (the $\BRT$ is inextensible and unshearable) is recorded.

\noindent
\textbf{Dimension reduction.}
A number of $N_\basis=3$ top components from PCA for each angular strain is selected with their coefficient distribution used for generating training data $\mathcal{X}$.

\noindent
\textbf{Neural network training.}
The input and output size of the neural network are $9N_\marker=27$ and $3N_\basis=9$. 
The number of neurons in the hidden layers are chosen to be 32 and 16.
The training and validation losses over 100 epochs for the $\BRT$ model are depicted in the right column of Fig.~\ref{fig:benchmark_training}(b). 
The losses decreased over two orders of magnitude and we terminate the training at the $100^\text{th}$ epoch.
The best out of fifty trained models is picked for the inference stage. 

\noindent
\textbf{Inference.}
The motion of bending, adding twist, bend releasing, and then twist releasing is being tested for reconstruction.
The motion cycle repeats three times to demonstrate the reproducibility.
The marker measurement set $\measurementset_t$ is received at 100 Hz from Vicon and real-time reconstruction is performed yielding the estimated strain $\hat{\strains}_t$ and reconstructed posture $\hat{\pose}_t$ results.
Fig.~\ref{fig:BR2_reconstruction_result}(a) demonstrates the trajectories of the ring marker at the tip of the $\BRT$ and the tip position of the reconstructed arm. 
The reconstruction error depicted in Fig.~\ref{fig:BR2_reconstruction_result}(b) shows that the reconstruction is consistent and remains accurate over experiments. 
The actual $\BRT$ posture and the reconstructed smooth arm are qualitatively compared at three time frames: pure bending, bending-twisting combination, and pure twisting, which are also shown in the estimated strain plots $\hat{\strains}_t(s)$ in Fig.~\ref{fig:BR2_reconstruction_result}.

\section{Conclusion and Future Work} \label{sec:conclusion}

In this paper, a neural network framework is proposed for real-time smooth posture reconstruction of a soft continuum arm robot. 
The challenge of learning continuous shape is addressed by dimension reduction through the principle component analysis. 
The training data for the neural network is generated by sampling from the distribution of strain coefficients, avoiding expensive and time-consuming data collection from experiments. 
The framework demonstrates reconstruction accuracy comparable to the benchmark algorithm while achieving computation speed over five orders of magnitude faster.
Future work will focus on extending the framework to handle more complex dynamic motions, such as bend propagation \cite{wang2022control}. 
An extension from a quasi-static to a dynamic setting is expected and requires integrating dynamic models for state estimation \cite{zheng2024estimating}.
Advanced machine learning techniques like LSTM and Transformers \cite{ding2021predictive, alkhodary2023learning} will improve the framework's ability to capture intricate soft arm dynamics.
These enhancements will be crucial for applying the framework to real-time sensory feedback control in soft robotic applications, improving its performance across a broader spectrum of motion and dynamic scenarios.

  \bibliographystyle{IEEEtran}
  \bibliography{reference}

\end{document}